# Generating Training Datasets for Legal Chatbots in Korean

**Hwang Chang-hoe,[1] Nam Jee-sun,[1] Eric Laporte [2]**
**[1] Hankuk University of Foreign Studies, DICORA, [2] Univ Gustave Eiffel, CNRS, LIGM**
**hch8357@naver.com, namjs@hufs.ac.kr, eric.laporte@univ-eiffel.fr**

***Abstract*— *Chatbots are robots that can communicate with humans using text or voice signals. Legal chatbots improve access to justice, since legal representation and legal advice by lawyers come with a high cost that excludes disadvantaged and vulnerable people. However, capturing the diversity of actual user input in datasets for deep-learning dialog systems (chatbots) is a technical challenge. Diversity requires large volumes of data, which must also be labelled in order to classify the user's intent, while the cost of labelling datasets increases with volume. Instead of labelling large volumes of authentic data from users, our approach consists in jointly generating large volumes of utterances and high-quality labels. The generator of labelled datasets is based on language resources that take the form of local grammar graphs (LGG), which capture and generalize the vocabulary and local syntax observed by linguists in text. The LGGs associate labels to the utterances according to a domain-specific classification system. We tested this approach by implementing LIGA, a legal chatbot in Korean. The chatbot answers users' conversational queries on legal situations by providing information on similar legal cases, made publicly available by the Korean government. We generated labelled utterances from the LGGs with the aid of the open-source Unitex platform. This process produced 700 million utterances. We trained a DIET classifier on a dataset made of these utterances, and the trained model reached 91% f1-score performance. We implemented a chatbot called LIGA, which uses the results of the model to select a link to a web page that documents similar legal cases.***

## I. Introduction

Modern society is based on the rule of law, and citizens are guaranteed the maximum freedom that they deserve through the law, while being controlled for actions that violate the public interest, and law is used as an active tool to resolve legal disputes that may arise between them. But even in the information society, there are many restrictions on using or receiving appropriate legal services. According to a survey of 3,839 men and women aged 19 to 75 in Korea in 2017, 26% of them had experience in needing legal services but not using them, and 47% of them did not use legal services because they were "difficult to approach." One of the reasons is that legal representation and legal advice by lawyers come with a high cost that excludes disadvantaged and vulnerable people. Legal chatbot systems, which are non-face-to-face legal advice services, are cheaper or free and therefore tend to improve access to justice [8]. Another obstacle is the wide gap between professional expressions used in the legal service field, and expressions actually used by the public in relation to legal disputes. As these issues are recognized throughout the legal industry, various legal chatbots have been launched to provide a connection to legal services based on expressions routinely used in relation to legal disputes.

In addition, many face-to-face services and activities have been converted to non-face-to-face format due to the COVID-19 pandemic, and the growth of related markets continues to be steep [1]. Such systems often include a service that connects a user and a professional, which generally requires natural language understanding (NLU) and chatbot technologies for automatically answering users' questions while limiting costs in manpower and resources.

This situation motivates research on chatbots and NLU. This study is conducted with the aim of analysing categories of law-related queries that cannot be analysed on existing online legal advice service sites, and of presenting a method of generating a dataset for training the NLU model of a chatbot system able to analyse queries for legal advice and provide specific information according to the relevant legal categories. This paper is organised as follows. The next section surveys related work. Section III reports the analysis of user queries, and Section IV describes the language resources built for this approach. Section V reports the experimentation carried out for evaluation, and Section VI contains concluding remarks.

## II. Related research

A representative example of an English-language legal chatbot is the Ross system developed by IBM. Ross is an 'AI lawyer' with 1 million pieces of advice from professional lawyers and machine learning algorithms. It identifies legal issues in sentences entered by users based on grammar structures, and outputs appropriate search results. Ross was tested for advice related to bankruptcy at BakerHostetler, one of the largest law firms in the United States, and reduced the working time to retrieve related laws by about 20% [6].

A representative example in Korean-language is Lawtalk, a service that connects lawyers and customers based on NLU technology, and provides search results by recognizing users' text related to legal issues. However, it has limitations. Let's look at the following examples:

(1) *The boss didn't pay me full salary and ran away. What should I do?*

(2) *I've been working at a convenience store for 4 months, but I'm quitting in about 2 months. Is there any way that the boss can report me to work longer than the contract says?*

First, the recognition of the meaning of the input text is limited: current systems cannot classify these queries as requesting legal advice on labour and wages. A second limitation is related to the length of the input sentence: (2) is too long to be processed.

Current research on chatbots addresses the NLU challenge, i.e. automatically analysing and classifying the speaker's intention, by using deep learning. This requires large training datasets. Given the considerable gap between legal terms used in ordinary language and in professional practice, these datasets must capture the diversity of actual user input, in everyday terms as well as in technical terms. However, collecting such data raises privacy issues, as most legal advice is related to personal history. In addition, these data must also be labelled in order to classify the user's intent, and the cost of labelling datasets increases with volume, making it the main bottleneck in data preprocessing tasks for machine learning [5]. Solutions such as weak supervision and crowdsourcing generate labels of uneven quality.

Instead of collecting large volumes of authentic data from users and labelling them, an alternative approach consists in jointly generating large volumes of texts and high-quality labels [9]. Since it is important to observe the linguistic aspects of authentic queries in order to carry out this task, we first collected anonymous legal advice query texts registered on a legal advice website. The lexico-syntactic patterns in the queries were formalised in the Local Grammar Graph (LGG) framework [3], which is useful for efficiently describing and processing local language phenomena into an exploitable NLU language resource. This resource was used to automatically generate a labelled dataset, which in turn trained an NLU model of chatbot.

## III. Analysis of authentic data

In order to analyse the patterns of expressions frequently input by actual users in queries for legal advice, 10,037 legal advice texts provided by the Korea Legal Aid Corporation and 4,586 queries for legal advice written by users of the legal advice service Lawtalk were collected, totalling almost 15,000 advice cases.

The next step consisted in establishing a legal classification system to serve as a framework for constructing language resources. We reviewed the systems of legal categories used in existing for-profit and non-profit legal services and we referenced the categories with the largest counts in the statistics of actual litigation cases provided by the courts of the Republic of Korea in the Judicial yearbook of 2020. We organized a legal classification system consisting of 20 categories organized in 4 larger categories: Divorce, Inheritance, Labour, and Protection of Personal Information.

For each of the 20 categories, characteristic expressions and keywords were examined in the collected data in a bottom-up manner. For example, the Divorce category is detailed in 5 subcategories: partner, child, family, cheater, and money, exemplified by the following representative queries:

(3) *I want to **divorce** from my **husband**.*

(4) *Can I **take** the **child** and **raise** him?*

(5) *Is it possible that my **mother-in-law** is a ground for divorce?*

(6) *Can I file for divorce on the **adultery** issue?*

(7) *Is it possible to claim **property division** in order to divide the apartment?*

Query (3) belongs to a category where verbs such as *divorce/break up* are associated with nouns such as *spouse*. The category of query (4) is characterized by nouns such as *child/daughter/son* and predicates related to parenting such as *take/raise*. (5) belongs to the category where expressions denoting relatives occur in queries as being a cause of divorce, and (6) to the category with expressions denoting third parties excluding family and spouse. Finally, in (7), an expression is related to property problems after the divorce.

Patterns of expressions related to each of the 20 categories were extracted from the actual text of the 15,000 advice cases in a data-driven manner and formally described in language resources.

The keywords classified in this taxonomy occur in the core of the queries. But many queries also describe the user's environment and situation:

(8) *I'**m in the middle of a divorce lawsuit** for cheating, namely **adultery by my husband**...*

(9) *I'm an office worker **who lost his wife in a car accident** recently and lives with younger siblings...*

In (8), *adultery by my husband* and *in the middle of a divorce lawsuit* explain the background of the consultation, and in (9), so does *lose wife in a car accident*. Such expressions are generally a description of events that occurred before the consultation. Background expressions were collected in the data, analysed, classified among the 4 categories, and formalized as linguistic patterns.

Finally, many queries contain a part that explicitly requests information:

(10) ***What** are visitation rights?*

(11) *Let me know **if** it's possible to report simple swear words as insult.*

(12) *Impersonation.*

Such expressions were classified in three categories: *who/what/where/when/why/how* questions as in (10), *yes/no* questions as in (11), and implicit requests as in (12), and the first two categories were formally described as patterns.

| [Background description] | [Core] | [Request of information] |
|---|---|---|
| - | 이혼 | 할 수 있나요 |
| 와이프 때문에 갈라서려는데 | 이혼시 필요한 서류 | 어떻게 되나요 |
| 남편이 바람을 피워서 | 이혼하려면 | 무엇을 해야하죠 |
| 배우자가 외도를 해서 | 이혼하는 방법이 | 알고 싶어요 |
| 남편과 갈라서고 싶은데 | 이혼을 위한 준비 | - |

Figure 1. Sample of target statements in the DIVORCE-PARTNER category

## IV. Building a language resource for NLU in the legal advice domain

The lexico-syntactic patterns inventoried and described during the analysis of the queries were used to build a language resource that represents the linguistic diversity of the queries. This resource is organised according to the LGG framework, an approach to language description that makes it easy to describe locally realized language patterns, such as multiword expressions, collocations and compositional expressions. It describes expressions in the form of directed acyclic graphs called LGGs, and the expressions described are compiled into a Finite-State Automata/Transducer (FSA/FST) format. The resulting resource can be used both in parsing-related tasks, to automatically index and annotate language expressions occurring in text, and in generation-related tasks, to generate in text format the expressions described in an LGG by enumerating all its paths. We used the Unitex platform [7] to construct, compile and use the LGGs.

The language patterns were built according to the Semi-automatic Symbolic Propagation (SSP) approach [4]. SSP is a method of building language resources on the basis of expressions occurring in text, for the purpose of generating annotated training data. The first step of SSP consists in formalizing patterns observed in annotated text. This bootstraps an iterative process where the resulting LGGs are applied to annotated text, and edited depending on the output. Finally, the LGGS are used to generate new annotated text. The approach allows for theoretically unlimited augmentation of the annotated data required for training.

A typical query for legal advice contains a background description, a core that indicates the user's intent, and an expression for request of information, as in Fig. 1 above. The LGG technology allowed for connecting these three parts in a modular way.

The language resources are organized in three parts in the same way. Four background expression modules are dedicated to each of the 4 categories in the legal classification, with an average of 97,861 expressions for each. The core modules are 20 subgraphs for the 20 subcategories, and each of these subgraphs contains an average of 1,724 paths. Two modules describe in their paths the expressions for request of information: one for *who/what/where/when/why/how* questions, with 1,109 paths, and the other for *yes/no* questions, with 766. The whole resource with all the modules totals about 700 million paths. Here is a sample of three utterances generated in the DIVORCE-PARTNER category:

(13) *I filed my marriage three years ago, and my wife and I are preparing for a divorce. I want to know what we should do to get a divorce.*

(14) *Teach me if you know what to do with divorce proceedings.*

(15) *I've been living with my wife since two years ago, and now I want to divorce her, so I want you to teach me how to file a divorce suit.*

The utterances were output with the aid of Unitex, by traversing the compiled resource, and data randomly extracted from them made up a training dataset.

As compared to building the same volume of NLU training data through crowdsourcing, the method proposed in this study is very advantageous in terms of cost and efficiency, but also as valid, systematic and significant as other research results based on the SSP approach [4].

## V. Checking the usefulness of the language resources

In order to assess the usefulness of the language resources proposed in this study, we used the training dataset to train an NLU model to classify the user's intent in queries for legal advice, we assessed the performance of the model and we used it in a legal chatbot.

The classifier was constructed with the aid of Rasa, [1] an open source framework for chatbot implementation. The Open Korean Tokenizer (OKT) was used to segment input text into morphemes. The architecture of the classifier is that of a dual intent and entity transformer (DIET) [2], an architecture designed to process named-entity recognition and intent classification at the same time, with better performance in terms of accuracy and speed than models trained separately for the two tasks. The model trained on the dataset selects items relevant to the user's intent as expressed in the input query, among the above-defined classification system.

The performance of the model trained on our dataset for the legal advice domain was evaluated on 200 sentences written by the operator. The results of the NLU analysis by

[1] https://rasa.com/

the DIET-based model are as follows:

| Precision | Recall | F1-score |
|---|---|---|
| 0.92 | 0.90 | 0.91 |

Here are two examples of atypical input queries where the model failed to predict the DIVORCE-PARTNER subcategory:

(16) *I've endured and lived a year since I found out that my pregnant wife met a man a year ago, but now I'm tired too.*

(17) *Can there be no alimony even though they threatened to rape and murder the baby?*

In (16), the core expressions characteristic of the category, i.e. directly or indirectly related to divorce, did not occur, and the absence of explicit expression of request of information did not help. In (17), the expressions of verbal abuse by the other party and threat of murder were not taken into account because they do not appear in the language resource. However, if the missing expressions are inserted in the resource, performance improvement can be expected after regeneration of the dataset and retraining. Therefore, it is possible to continuously work for optimizing performance by inventorying error types, re-analysing relevant expressions, and reflecting them in the language resources.

The performance of the NLU model trained through the above process allowed for implementing a legal advice chatbot called LIGA (LInito leGal Assistant [2]). LIGA expects users' queries for legal advice on his or her situation, analyses them, classifies the situation and provides appropriate advice in the form of a link to a web page with similar legal cases. Representative legal cases were selected among the data disclosed in the case book of the Korea Legal Aid Corporation. Thus, queries such as *I want to sue him because I know that my personal information has been leaked without permission* and *The company does not comply with the working hours written in the contract* produce information about related legal cases.

This suggests the possibility that the language resources proposed in this study can be used in the actual application field, and that generation of NLU training data through the construction of language resources can be applied to other domains.

## VI. Conclusion

The approach tested in this study analyses the legal expressions occurring in discourses actually used for legal advice, constructs language resources by describing them in the LGG format, and uses them to generate an annotated training dataset for the corresponding domain. The classifier trained on the NLU dataset showed a high f1-score of 91%. A legal advice chatbot was constructed with the classifier. Its actual operation is more public-friendly than existing systems, since it goes beyond simple keyword recognition, uses a classification based on actual expressions used in authentic queries instead of technical terms, and takes into account the description of the context of the queries. This confirmed the usefulness of the approach and language resources. In the future, it is expected that the approach can be used to implement NLU models and chatbots in other domains.

[2] http://linito.kr/liga_legal_advisor_chatbot/